\documentclass[10pt,twocolumn,letterpaper]{article}

\usepackage{3dv}
\usepackage{times}
\usepackage{epsfig}
\usepackage{graphicx}
\usepackage{amsmath}
\usepackage{amssymb}
\usepackage{authblk}

\usepackage{soul}

\newcommand{\bmethodlabel}[2]{\scriptsize {\hspace{#1} {#2}}}

\usepackage[pagebackref=true,breaklinks=true,colorlinks,bookmarks=false]{hyperref}

\threedvfinalcopy 

\def\threedvPaperID{280} 

\ifthreedvfinal\fi
\begin{document}

\title{Deep Physics-aware Inference of Cloth Deformation for Monocular Human Performance Capture}

\author[1,2,3]{\vspace{-8mm}Yue Li}
\author[1,2]{Marc Habermann}
\author[3]{Bernhard Thomaszewski}
\author[3]{Stelian Coros}
\author[4]{Thabo Beeler}
\author[1,2]{Christian Theobalt}

\affil[ ]{
\vspace{-5mm}\textsuperscript{1}Max Planck Institute for Informatics,
\textsuperscript{2}Saarland Informatics Campus,
\textsuperscript{3}ETH Zurich,
\textsuperscript{4}Google Inc.
}

\maketitle
\thispagestyle{empty}

\begin{abstract}
Recent monocular human performance capture approaches have shown compelling dense tracking results of the full body from a single RGB camera.
However, existing methods either do not estimate clothing at all or model cloth deformation with simple geometric priors instead of taking into account the underlying physical principles.
This leads to noticeable artifacts in their reconstructions, \eg baked-in wrinkles, implausible deformations that seemingly defy gravity, and intersections between cloth and body.
To address these problems, we propose a person-specific, learning-based method that integrates a
simulation layer into the training process to provide for the first time physics supervision in the context of weakly supervised deep monocular human performance capture.
We show how integrating physics into the training process improves the learned cloth deformations, allows modeling clothing as a separate piece of geometry, and largely reduces cloth-body intersections. 
Relying only on weak 2D multi-view supervision during training, our approach leads to a significant improvement over current state-of-the-art methods and is thus a clear step towards realistic monocular capture of the entire deforming surface of a clothed human.

\end{abstract}

\begin{figure}[t]
    \centering
    \includegraphics[width=0.45\textwidth]{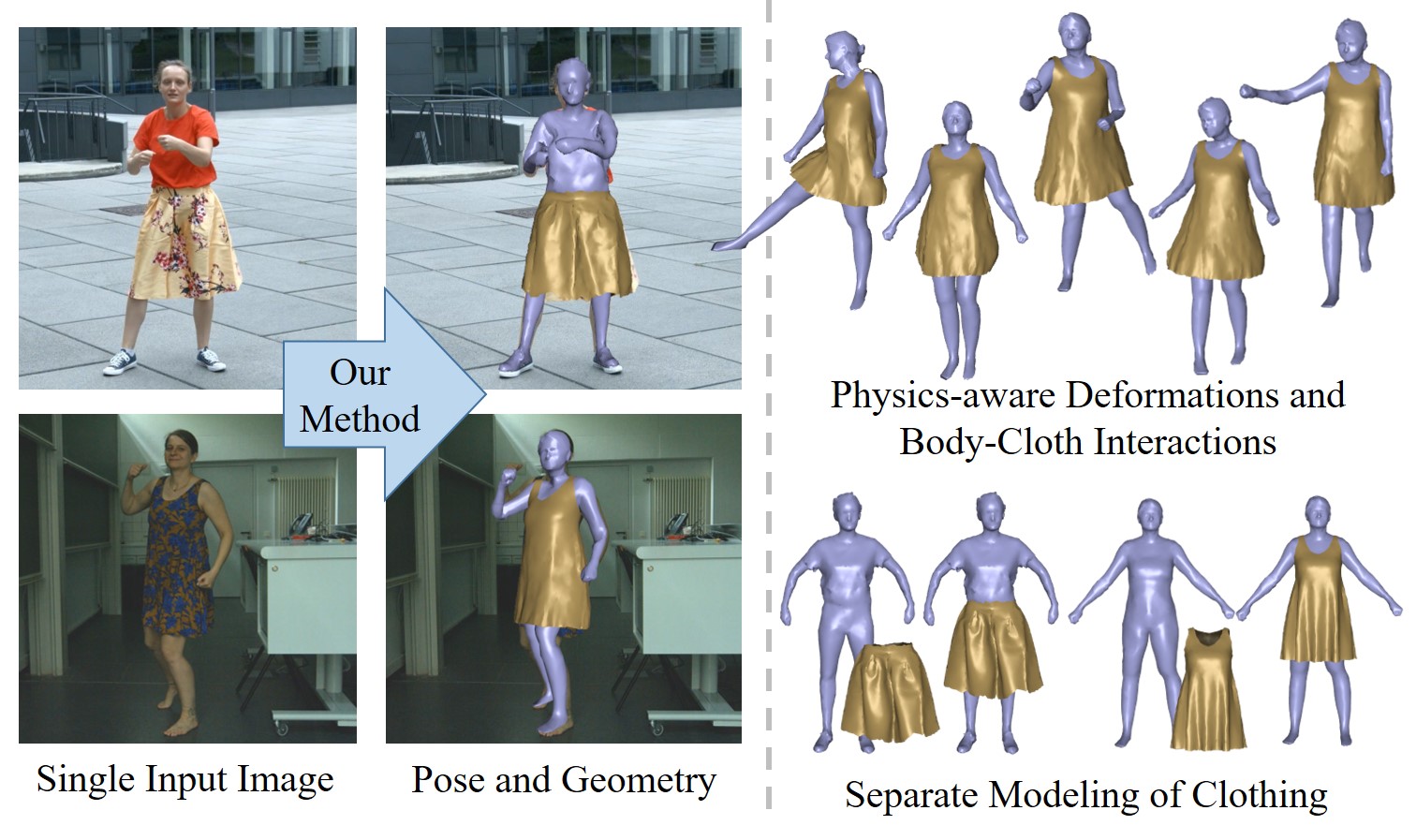}
    \caption{
    Our method estimates body pose and physically plausible surface deformation from a single image. 
    Importantly, body and clothing are represented as separate meshes allowing for accurate modeling of body-cloth interactions.
    }
    \label{fig:teaser}
\end{figure}

%
%
\section{Introduction}
\label{sec:intro}
Human performance capture plays a critical role in various computer graphics and vision applications such as virtual try-on, movies as well as video games.
With rapid progress in display and capture technology, expectations on the quality of geometric reconstruction and tracking are constantly increasing. 
Here, not only the geometric details are of major importance but also that the deformed and posed reconstructions follow the physical behavior of real objects which includes realistic wrinkle patterns as well as coherent interaction of body and clothing.
While professional content production studios can rely on involved multi-camera setups to capture high-fidelity human performances, there is an ever-growing desire to democratize performance capture for everyday applications, \eg virtual try-on, by utilizing much simpler and cheaper capture devices.
%
\par
Hence, research has shifted from expensive and complex multi-view capture setups~\cite{pons15,bray06,cagniart10,brox10,pons17,de08,gall09,liu11,mustafa15,vlasic09, wu13,pons17} to depth cameras~\cite{slavcheva17,guo17,newcombe15, innmann2016,guo18, kowdle18,ye12, dou16, yu18, ye14,wei12,yu2019simulcap} over the past decade. 
Unfortunately, the latter are sensitive to bright sunlight and thus are not suited for outdoor use-cases.
In conjunction with the advances in deep learning, the most recent research has shifted its attention onto \textit{single RGB camera} setups, offering the most flexible and low-cost setup.
Previous monocular methods have made a substantial progress in recovering the 3D unclothed body~\cite{hmrKanazawa17,pavlakos2018learning,kanazawa2019learning}, hand pose~\cite{wang_SIGAsia2020,GANeratedHands_CVPR2018,zhou2019monocular}, facial identity and expression~\cite{kim2018deep,tewari2017self,tewari17MoFA} as well as jointly tracking all of those~\cite{pavlakos2019expressive,xiang2019monocular,joo2018total, zhou2020monocular}.
However, only a few methods~\cite{habermann2020deepcap,habermann2019livecap,xu2018monoperfcap} coherently track the dense surface deformations with clothing included from monocular views, which is essential for a majority of applications.
These person-specific methods densely deform and pose a geometry to match the body pose and the clothing deformation in the input image while assuming an initial template of the person is given.
Recent learning-based monocular methods~\cite{habermann2020deepcap} only leverage image-based supervision, rendering it a challenging task to densely supervise deformations.
This manifests in simplified model assumptions, \eg, a single geometry for both body and clothing, or the utilization of simple geometric priors disobeying physics principles.
Consequently, they either fail to account for body-cloth interactions or having deformations that do not follow physical rules. Importantly, artifacts such as baked-in wrinkles from the initial scan are highly noticeable in their results.
%
\par
To this end, we propose a learning-based approach for capturing the body pose and the physically plausible clothing deformation from a single RGB image (see Fig.~\ref{fig:teaser}).
Our method comprises two networks dedicated to regress body pose in terms of joint angles and surface deformations in form of embedded deformation.
Importantly, during training, we only assume weak supervision with multi-view imagery, \ie 2D skeletal joint detection, and foreground masks.
These supervisions alone can hardly ensure physically plausible results.
Thus, at the core of our approach, we propose an efficient simulation layer that for the first time allows physically plausible self-supervision \textit{during} the training in such a weakly supervised setting.
We achieve this by integrating a physics-based simulator into a learning architecture that takes intermediate predictions of cloth and body positions and velocities to perform forward simulations.
The simulation results are then used to supervise the cloth deformations during training.
As cloth-body collisions are explicitly handled in the proposed layer, we can accurately model clothing as a separate piece of geometry in contrast to previous monocular methods.
%
%
%
In summary, our contributions are:
\begin{itemize}
\item A monocular human performance capture approach, which outputs body pose and physically plausible cloth deformations for dressed subjects.
\item A simulation network layer that allows on-the-fly simulation supervision \textit{during} training, which also enables separate modeling of cloth and body geometry.
\end{itemize}
In contrast to prior work, our method reconstructs physically more accurate deformations without baked-in wrinkles and with correct body-cloth collision handling. 
Our quantitative evaluations indicate that incorporating physics-based simulation during training provides significant improvements over state-of-the-art methods.
%
%
%
\section{Related Work}
\label{sec:relatedwork}
%
%
As our goal is recovering a dense surface of the human, we focus on previous works that achieve this by using parametric body models or template meshes, and works that treat body and clothing as separate mesh layers.
We omit the works on 2D~\cite{cao18,cao17,simon17,wei16} and 3D skeletal pose estimation~\cite{VNect_SIGGRAPH2017,XNect_SIGGRAPH2020,inthewild3d_2019,rhodin2018learning,popa2017deep,zhou2017towards,sun2017compositional,tome2017lifting,rogez_lcr_cvpr17} as they are not concerned with the problem of surface reconstruction.
%
%
\paragraph{Reconstruction of Parametric Body Models.}
The works~\cite{zhou2010parametric,jain2010movie,rogge2014garment,guan2009estimating, bogo2016smpl,Lassner,hmrKanazawa17,varol18_bodynet} that fall into this category use parametric body models~\cite{loper2015smpl}.
Some works fit the model parameters to sparse 2D and 3D joint predictions~\cite{bogo2016smpl} or regressed vertex positions~\cite{kolotouros2019convolutional} by minimizing corresponding energies.
Others~\cite{hmrKanazawa17} directly regress these parameters from images.
A set of recent works~\cite{pavlakos2019expressive,xiang2019monocular} extended body models to account for varying hand poses and facial expressions to jointly capture hands, face, and body.
While motion and shape of the undressed body are reconstructed, clothing is not considered.
%
%
\paragraph{Unified Reconstruction.}
One stream of previous work treats body and clothing as a single geometry.
Volumetric representations~\cite{zheng2019deephuman,varol2018bodynet} use an occupancy grid to represent the body, meaning that the resolution is limited by the grid.
Implicit methods~\cite{saito2019pifu,huang2020arch,saito2020pifuhd} methods overcome this limitation by treating the surface as an implicit function.
However, both approaches require post-processing to recover explicit surface representations.
Lacking temporal consistency thus prohibits these approaches for applications such as texture replacement or motion retargeting. 
Closely related to our work are template-based methods~\cite{habermann2019livecap, habermann2020deepcap, xu2018monoperfcap,carranza2003free, de2008performance} that track a template based on image observations.
Using a mesh with fixed topology as a reference, surface correspondence over time is explicitly given.
With input data originating from images only, this ill-posed problem is countered by simplified assumptions and geometric priors.
Consequently, static wrinkles contained in a template remain visible across all poses, and deformations commonly appear to be physically implausible, \eg defying gravity.
Most importantly, all these methods treat clothing and body as a single piece of geometry ignoring dynamic body-cloth interactions.
To address these limitations, we propose a simulation layer that encourages cloth deformations to not only satisfy image constraints but also exhibit physically plausible behavior.
%
%
\paragraph{Cloth as a Separate Part.}
In contrast to the above methods, there is also a line of work that reconstructs body and clothing as separate geometries.
Bhatnagar \etal~\cite{bhatnagar2019mgn} recover static geometry for clothing and body from a set of RGB images.
DeepWrinkles~\cite{lahner2018deepwrinkles} enables posing a piece of cloth where their method learns to regress pose-dependent wrinkles at high resolution.
ClothCap~\cite{pons2017clothcap} uses multi-view capture to produce a clothed human body that can be used for re-targeting. 
Stoll \etal~\cite{stoll2010video} recover cloth material parameters from multi-view video sequences to reproduce the observed garment deformation.
SimulCap~\cite{yu2019simulcap} performs quasi-static physics simulation with depth matching constraints to reconstruct the clothing layer.
Different from the above methods, our approach relies solely on a monocular RGB camera.
Also leveraging simulation, MulayCap~\cite{su2020mulaycap} recovers, both, the texture and the geometry of a dressed subject from monocular RGB videos by a multi-layer decomposition approach. 
However, simulation is only used to generate initialization for the succeeding refinement stage, whereas we consistently enforce simulation supervision.
\par
As a potential alternative to simulation, geometric detail such as wrinkles can be added in a data-driven, pose-dependent manner \cite{guan2012drape,santesteban2019learning,patel2020tailornet,gundogdu2019garnet}.
Different from these geometry-driven methods, we integrate physics-based simulation into our training framework thus encouraging physical plausibility with only a single image as input.
%
%

%
%
\section{Method}
\label{sec:method}
\begin{figure*}[ht]
    \centering
    \includegraphics[width=\textwidth]{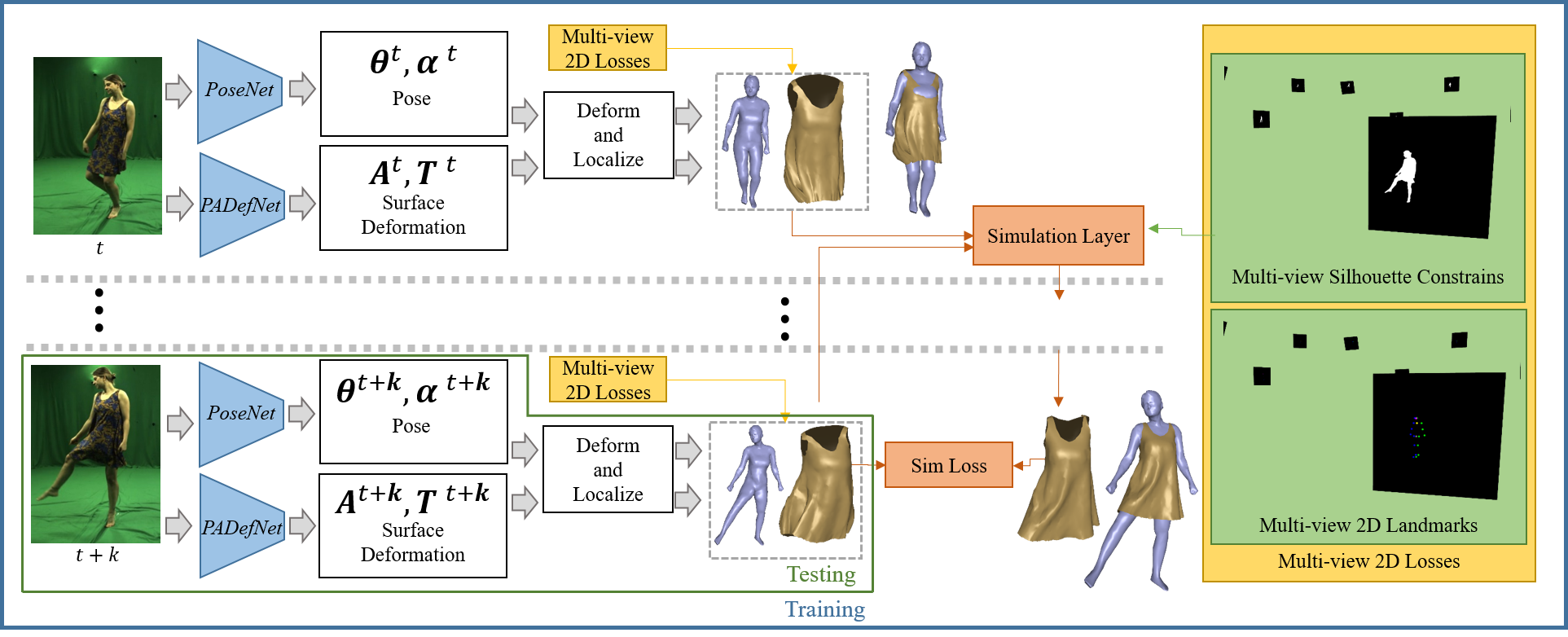}
    \caption{
    Our method takes a single image as input and two networks, \textit{PoseNet} and \textit{PADefNet}, regress the skeletal pose as well as embedded deformation parameters for the clothing.
    Combining the outputs of the two networks allows posing and deforming the body and clothing geometry.
    During training, we use multi-view image losses for \textit{PoseNet} and \textit{PADefNet} is additionally supervised by our proposed simulation loss to encourage physically plausible deformations.
    To evaluate the simulation loss, we run on-the-fly cloth simulation on small windows of subsequent frames from the training sequence and penalize the difference between regressed deformations and simulation outputs.
    }
    \label{fig:pipeline}
\end{figure*}

Our template-based method leverages a deep neural architecture, taking a single background-segmented person image as input and regresses posed and deformed surface meshes for body and clothing which match the performance in the input image (Fig.~\ref{fig:pipeline}).
Before training, a 3D template of the person with separate cloth and body geometry and a multi-view recording of the subject performing various motions has to be acquired (Sec.~\ref{sec:method_data}). 
The technical core of our architecture is formed by two prediction networks, \textit{PoseNet} and \textit{PADefNet}, that are trained to regress body pose and physics-aware cloth deformation, respectively (Sec.~\ref{sec:method_surface_prediction}).
\textit{PoseNet}~\cite{habermann2020deepcap} regresses skeleton joint angles and the root rotation from the input image using multi-view 2D joint detection as weak supervision.
The proposed \textit{PADefNet} predicts the surface deformation of the cloth template by regressing embedded graph parameters from the same input image. 
In addition to multi-view image data, \textit{PADefNet} leverages our cloth simulation layer as supervision, which encourages physically plausible deformations (Sec.~\ref{sec:method_sim}).
%
%
%
%
%
\subsection{Data Processing}
\label{sec:method_data}
%
%
\paragraph{Template Acquisition.}
\label{sec:tempalte_acquisition}
Similar to DeepCap\cite{habermann2020deepcap}, we acquire a single scan for body and clothing (\eg using photogrammetric scanning). 
Surface registration against a parametric body mesh model 
\cite{alldieck19cvpr,bhatnagar2020ipnet} is then performed to obtain an estimate for body parts occluded by clothing, \eg the legs under a skirt, which are merged with the visible body parts from the scan to form a complete body mesh.
The arms of the body mesh are labeled as inactive when resolving collisions.
A separate cloth mesh is created manually from the scan, a task could also be automated~\cite{stoll2010video,pons2017clothcap}.
Skeleton parameters and skinning weights, required for posing the meshes, are determined automatically~\cite{habermann2020deepcap}. 
Two separate embedded graphs~\cite{sumner2007embedded,sorkine2007rigid} for body and clothing are computed by down-sampling the original meshes.
These pre-processing steps only need to be done once per character.
For more details, we refer to the supplemental document.
%
%
\paragraph{Video Capture.}
We capture the subject to be tracked in a multi-view green screen studio with calibrated and synchronized cameras.
The person is asked to perform various tasks, \eg walking, and dancing, to best sample the space of possible poses.
Next, we apply OpenPose~\cite{cao17,cao18,simon17,wei16} on all frames and views to obtain multi-view 2D joint predictions.
Color keying is used to segment the foreground from the green-screen background and compute distance transformation images $\boldsymbol{D_c}$ from the foreground masks~\cite{borgefors86}.
%
%
\subsection{Cloth Simulation Layer}
\label{sec:method_sim}
Our simulation layer uses the publicly available cloth simulation framework ARCSim~\cite{narain2012adaptive} as its basis, but we make several adjustments that we describe below. 
%
%
\paragraph{Material Model and Parameter Selection.} 
ARCSim leverages a data-driven material model defined by a total of 39 parameters. 
While parameter values for several real-world fabrics are provided, we found that none of them were ideally suited for the materials that we use in our examples, and manually adjusting parameters to obtain better approximation proved very difficult. 
For this reason, we resorted to a simpler, isotropic material model\cite{thomaszewski2008asynchronous} defined through three parameters: 
Young's modulus and Poisson's ratio for in-plane behavior, and a single bending stiffness coefficient. 
We determine parameter values through best-guess initialization and a few iterations of simulation-based tuning to better approximate the qualitative behavior observed in the input video sequences. 
We used the same parameters for all sequences. 
Although this manual approach is sufficient for our examples, this task could be further automated~\cite{stoll2010video}.
%
%
\paragraph{Time Integration.} 
During training, the initial state and velocities that are fed into the simulation layer can exhibit large deformations that, when using ARCSim's default integration method, can lead to instabilities.
To improve stability, we resort to an optimization-based formulation of fully implicit Euler~\cite{Martin10EBEM} combined with adaptive regularization and a back-tracking line search.
Finally, we make several code adaptations to enable batch operations for efficient training and integrate the simulation engine in a customized TensorFlow\footnote{https://www.tensorflow.org/} layer.
We refer to this layer as the simulation function $\mathcal{S}$, which takes cloth and body vertices as input and returns the cloth positions for the next time step.
%
%
\paragraph{Silhouette Constraint.}
While our simulation model captures the characteristic behavior of clothing, it is still an approximation, and deviations from the input images must be expected due to external forces such as air drag, viscous damping, and friction that are not modeled. 
To better track the real-world behavior, we add a multi-view silhouette constraint term.
Specifically, this constraint ensures that the vertices $\boldsymbol{\Tilde{V}}_{\mathrm{cloth}}^t$ of the simulated cloth geometry matches the image silhouettes from all camera views for frame $t$. 
We construct a 3D ray going through the camera origin and the silhouette pixel $p$ and search for the boundary vertex $\boldsymbol{\Tilde{V}}_{\mathrm{cloth},p}^t$ that minimizes the distance to this ray.
The closest point on the ray is used as 3D point correspondence for the boundary vertex, enforced via soft constraints
%
%
\begin{equation}
    E_{cons} =\sum_p ||\boldsymbol{\Tilde{V}}_{\mathrm{cloth},p}^t - \boldsymbol{V}_{\mathrm{ray},p}^t||^2 .
\end{equation}
%
%
%
\subsection{Pose and Deformation Regression}
\label{sec:method_surface_prediction}
We separate the task of regressing the full surface deformation into predicting pose and surface deformation independently.
Therefore, our method consists of two ResNet50 based CNNs~\cite{he2016deep}, \textit{PoseNet} and \textit{PADefNet}, which regress skeleton pose and embedded deformation parameters from a segmented input image, respectively. 
%
%
\subsubsection{Pose Regression and Deformation Model}
To pose and deform template vertices as well as sparse body markers, a deformation layer~\cite{habermann2020deepcap} denoted as
%
%
\begin{equation}
\label{eq:posedeformationmodel}
    \boldsymbol{V}_{loc},\boldsymbol{K}_{loc} = f(\boldsymbol{\theta}, \boldsymbol{\alpha}, \boldsymbol{A}, \boldsymbol{T})
\end{equation}
%
%
is used, which is a combination of dual quaternion skinning~\cite{kavan07} and embedded deformation~\cite{sumner2007embedded,sorkine2007rigid}.
It takes the pose in terms of skeleton joint angles $\boldsymbol{\theta} \in \mathbb{R}^{3}$ and camera-relative root joint rotation $\boldsymbol{\alpha}\in \mathbb{R}^{3}$ as well as the embedded graph node rotation $\boldsymbol{A} \in \mathbb{R}^{K \times 3}$ and translation $\boldsymbol{T} \in \mathbb{R}^{K \times 3}$ where each row encodes rotations in terms of Euler angles and translation vectors for each of the $K$ nodes.
The output is the posed and deformed vertices $\boldsymbol{V}_{loc}$ and markers $\boldsymbol{K}_{loc}$ in camera and root-relative space 
.
The body pose parameters $\boldsymbol{\theta}$, $\boldsymbol{T}$ and $\boldsymbol{\alpha}$ are obtained from \textit{PoseNet}~\cite{habermann2020deepcap}.
%
%
\subsubsection{Physics-aware Deformation Regression}
To not only pose the template but also account for surface deformation, a dedicated network \textit{PADefNet} predicts the translation vectors $\mathbf{T}$ and rotation angles $\mathbf{A}$ of the embedded graph (EG) from the segmented input image. 
\textit{PADefNet} is supervised using a combination of both image-based and physics-based metrics, which ensure that deformations match image-based observations while minimizing violations of physical equilibrium conditions.
In the remainder of this chapter, we assume \textit{PoseNet} is fixed and provides the posed and deformed vertices $\boldsymbol{V}$ and markers $\boldsymbol{K}$ in global space.
As $\boldsymbol{V}$ and $\boldsymbol{K}$ are a function of the \textit{PADefNet} outputs ($\mathbf{T}$, $\mathbf{A}$), we can then supervise \textit{PADefNet} on $\boldsymbol{V}$ and $\boldsymbol{K}$.
%
%
\paragraph{Warm Start.}
To jump-start our training including the simulation layer, we first pre-train \textit{PADefNet} without running simulation but use a geometric regularizer (ARAP~\cite{sorkine2007rigid}). 
This adds robustness to the training as geometric regularizers are more stable than simulation and significantly reduces overall training time. Once the network predicts reasonable shapes, we add the simulation loss to supervise the physical deformation.
The loss is defined as
%
%
\begin{equation}
   L_{\mathrm{warm}} = L_{\mathrm{sil}} + L_{\mathrm{lm}} + L_{\mathrm{reg}} + L_{\mathrm{att}}
\end{equation}
which comprises multi-view losses as well as geometric priors. The individual loss terms are defined as follows.
%
%
\paragraph{Multi-view Losses.}
Our multi-view 2D landmark loss
%
%
\begin{equation}
\label{lm_loss}
    L_{\mathrm{lm}} = \beta_{\mathrm{lm}} \sum_c \sum_{m} ||\Pi_c(\boldsymbol{K}_{m}) - \boldsymbol{p}_{c,m}||^2
\end{equation}
%
%
ensures that the projected landmark matches the 2D detection $\boldsymbol{p}_{c,m}$ for all views $c$ and landmarks $m$.
Here, $\Pi_c$ denotes the projection function of the camera $c$.
To densely supervise the surface, we also introduce a silhouette loss
\begin{equation}
\label{sil_loss}
     L_{\mathrm{sil}} = \beta_{\mathrm{sil}} \sum_c \sum_{b\in \mathcal{B}_c} \rho_{c, b} ||\Pi_c(\boldsymbol{V}_b) - \boldsymbol{D_c}||^2,
\end{equation}
which ensures that the set of mesh boundary vertices $\mathcal{B}_c$ matches the zero contour line in the distance transformation image $\boldsymbol{D_c}$ for all views.
$\rho_{c, b}$ is a weighting term ensuring that silhouettes are only matched if the normal of the surface aligns with the gradient of the distance transformation~\cite{habermann2019livecap}.
%
%
\paragraph{Regularization Loss.}
To regularize deformations and to avoid drifting of the surface, we employ the as-rigid-as-possible prior~\cite{sorkine2007rigid} to ensure smooth local embedded deformations.
We further adopt the rigidity weights formulation~\cite{habermann2020deepcap} to model material-dependent deformation behaviors, \eg the skirt can deform more freely than the skin.
%
%
\paragraph{Attachment Loss.}
Note that our entire mesh $\boldsymbol{V}$ can be split into body and garment meshes, denoted as $\boldsymbol{V}_{\mathrm{cloth}}$ and $\boldsymbol{V}_{\mathrm{body}}$ in the remainder of this section.
To ensure a coherent movement of these two, an attachment loss 
%
%
\begin{equation}
\label{att_loss}
    L_\mathrm{att} = \beta_\mathrm{att} \sum_{i \in \mathcal{A}}  ||\boldsymbol{V}_{\mathrm{cloth},i} - \sum_{j=0}^2 \gamma_{i,j} \mathcal{C}(\boldsymbol{V}_{\mathrm{cloth},i},\boldsymbol{V}_{\mathrm{body}})_j||^2
\end{equation}
%
%
is included to ensure that the cloth is attached to the body at some anchor positions, \eg the waistband of a skirt has to be attached to the hip of the body mesh.
Here, $\mathcal{A}$ are the selected vertices on the garment that act as anchor points, $\mathcal{C}$ is a function that takes the cloth vertex id $i$ and returns the 3 vertices of its closest triangle on the undeformed body mesh, and $\gamma_{ij}$ are barycentric weights computed from the closest point on this triangle and its three vertices.
%
%
\paragraph{Physics-aware Training.}
\label{sec:method_phys_network}
While the previous training stage constrains the surface mesh to match the image evidence, it can neither account for the collision of body and clothing nor ensure physically plausible cloth deformations.
To this end, we introduce a dedicated simulation-based loss as a better substitution for the ARAP term to explicitly penalize collision behavior and physically implausible deformations.
Our final loss is then defined as
%
%
\begin{equation}
    L = L_{\mathrm{sil}} + L_{\mathrm{lm}}  + L_{\mathrm{sim}} + L_{\mathrm{att}}.
\end{equation}
%
%
\par
As our simulation layer $\mathcal{S}$ is directly integrated into a learning framework, we can perform on-the-fly simulation during training.
While our method takes a single image as input, the simulation-based loss term is designed to be a multi-frame function to better leverage the sequential training data available.
More concretely, this term penalizes the accumulated error on a set of consecutive frames, \ie the mismatching between the per-frame predictions and the on-the-fly simulation results within a frame window (see Eqn.~\ref{eqn:sim_loss}).
We found that in practice performing simulation over long sequences is extremely challenging when using shapes and poses predicted by a network, since even visually unnoticeable errors, \eg cloth getting trapped in body self-intersections, can lead to catastrophic failures. 
Hence, we designed our framework specifically to rely only on small simulation windows $\mathcal{F}$ starting at random frames $t'$ to have shorter but successful simulations for training.
Additionally, the chosen design is well suited for machine learning, as it allows to access data randomly and in parallel for training.
%
%
In the following, we refer to a specific frame in this window using the superscript $\cdot^t$, where $t \in \{t', ..., t'+ \mathcal{F} \}$.
Our physics loss then reads
%
%
\begin{equation}
\label{eqn:sim_loss}
    L_\mathrm{sim} = \beta_\mathrm{sim} \sum_{i} \sum_{t=t'+1}^{t'+ \mathcal{F}} ||\boldsymbol{V}_{\mathrm{cloth},i}^{t} - \boldsymbol{\Tilde{V}}_{\mathrm{cloth},i}^{t}||^2.
\end{equation}
%
%
%
Here, $\boldsymbol{\Tilde{V}}_{\mathrm{cloth},i}^{t}$ denote the post-simulation cloth vertex positions, defined as
%
%
$$
\boldsymbol{\Tilde{V}}_{\mathrm{cloth}}^{t} = 
   \begin{cases}
        \boldsymbol{V}_{\mathrm{cloth}}^{t}, & t = t'  \\
        \mathcal{S}( \boldsymbol{V}_{\mathrm{cloth}}^{t-1}, \boldsymbol{V}_{\mathrm{cloth}}^{t}, \boldsymbol{V}_{\mathrm{body}}^{t-1},  \boldsymbol{V}_{\mathrm{body}}^{t}),              & t = t'+1\\
        \mathcal{S}(\boldsymbol{\Tilde{V}}_{\mathrm{cloth}}^{t-2}, \boldsymbol{\Tilde{V}}_{\mathrm{cloth}}^{t-1}, \boldsymbol{V}_{\mathrm{body}}^{t-1},  \boldsymbol{V}_{\mathrm{body}}^{t}), & t > t'+1
    \end{cases}
$$
%
%
where $\mathcal{S}$ is the aforementioned simulation operation.
We initialize the cloth position with \textit{PADefNet} outputs at $t = t'$, where no history is available, and the velocity is initialized using finite difference with cloth position between the succeeding frame.
The body vertices positions come from the network predictions and velocities are computed always with finite differences.
The loss is then evaluated on the $\mathcal{F} - 1$ frames (excluding the first frame). 
Even though the first frame in a training sample sequence does not receive this supervision, that frame is supervised by our multi-view supervision, such that in practice all frames are supervised.
We opted to not backpropagate gradients through the simulation inputs with respect to the EG parameters as guaranteeing convergence during training would be harder. 
%

%
%
\section{Results}
\label{results}
\begin{figure*}[t]
    \centering
    \includegraphics[width=\textwidth]{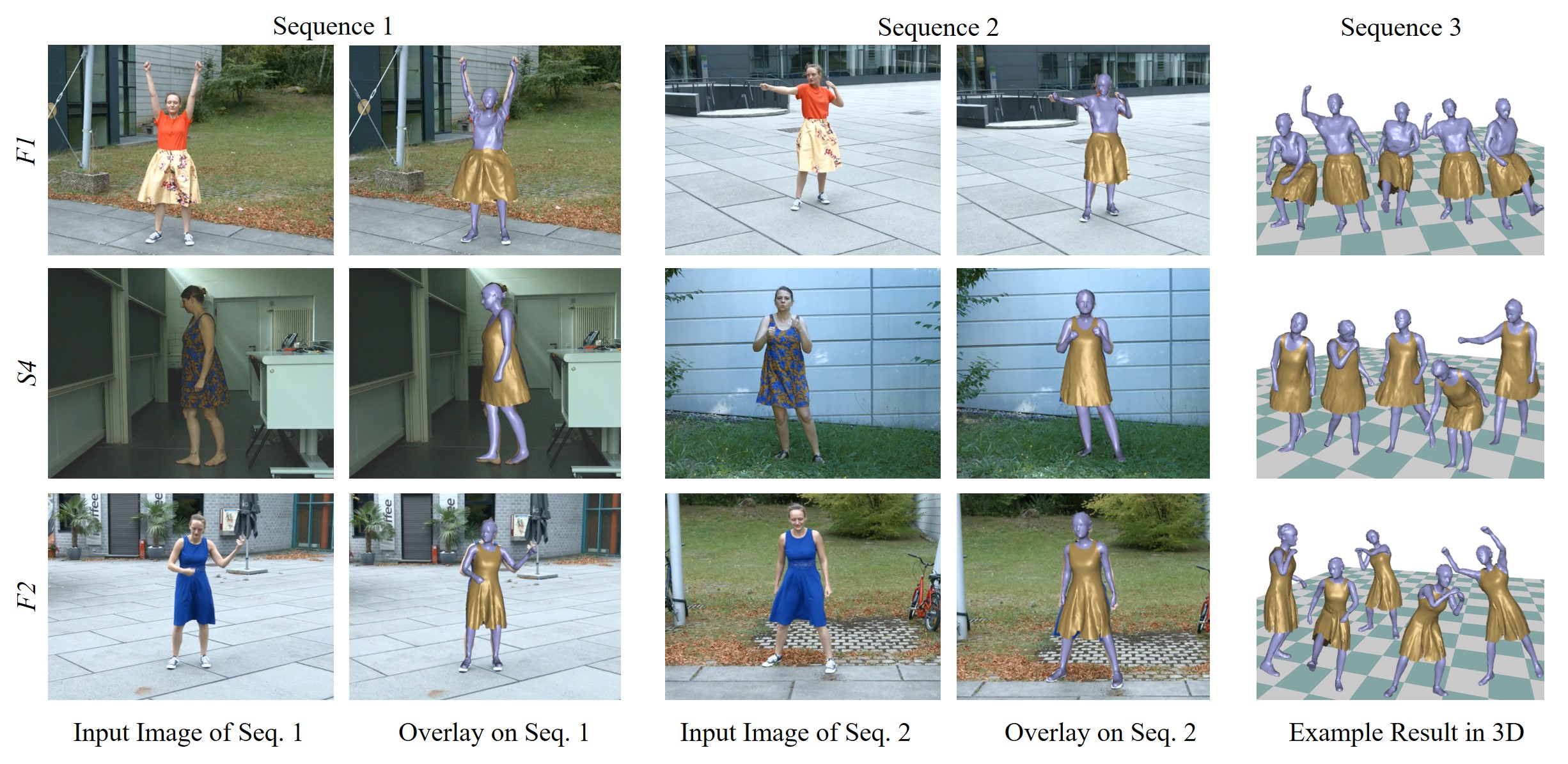}
    \caption
    {
    Reconstructions obtained with our method for various in-the-wild environments and challenging motion combinations. 
    Our results show good overlay quality throughout, attesting to the pose and clothing estimation accuracy of our method. Furthermore, diverse and physically plausible cloth deformations are observed for a wider range of poses.
    }
    \label{fig:our_results}
\end{figure*}

\begin{figure}[t]
    \centering
    \includegraphics[width=\linewidth]{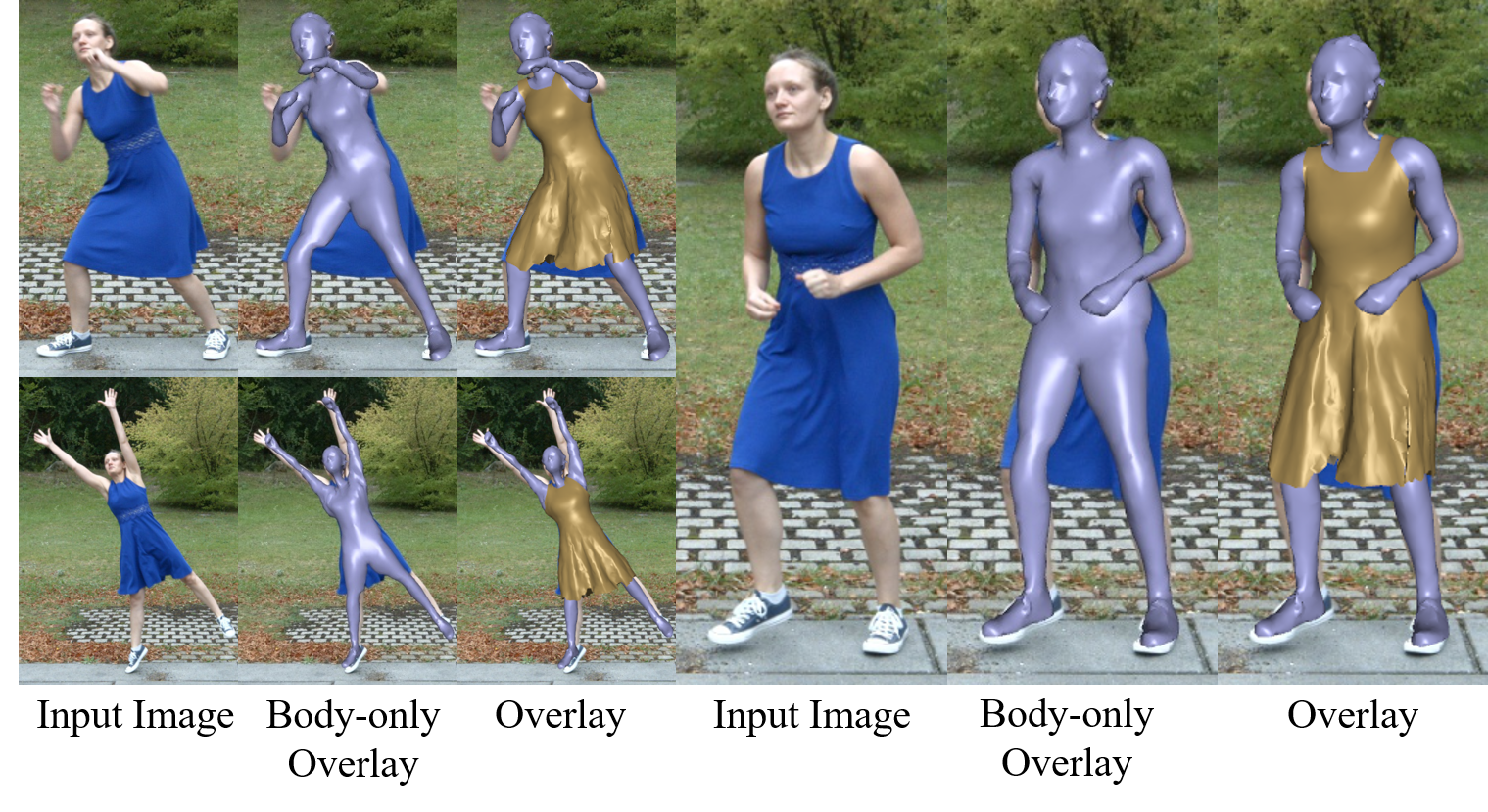}
    \caption{
    Reconstruction results.
    Despite minor distortions in occluded regions, our reconstructed body geometry matches the image evidence. 
    Thanks to the separate modeling, our cloth reconstruction is able to reproduce the folding and unfolding behavior of the dress driven by the underlying body motions.
    }
    \vspace{-3mm}
    \label{fig:dynamic_wrinkle}
\end{figure}

We evaluate our approach on various outdoor and indoor environment settings with three subject-cloth combinations under a wide range of motions (see Fig.~\ref{fig:our_results}).
To bridge the domain gap between training data recorded in the capture studio and in-the-wild testing sequences, \eg different light conditions, we apply a domain adaptation step.
\textit{PoseNet} and \textit{PADefNet} are refined for 300 iterations on the testing sequence leveraging the losses introduced before but \textit{using only a single camera}. 
For in-the-wild captures with varying and dynamic backgrounds, we segment the input images using OSVOS~\cite{caelles17}.
While the result is almost collision-free thanks to \textit{PADefNet}, minute intersections can remain, which is why we run a final collision resolution step (see also supplemental video). 
This optional step takes 2s per frame on an Intel i7-9700 CPU.
Following DeepCap~\cite{habermann2020deepcap}, we apply a temporal Gaussian filter of size 5 frames.
%
%
\vspace{-10pt}
\paragraph{Dataset.}
Our training dataset contains 3 green screen studio capture sequences with actors performing a large range of motions.
For testing, we recorded an additional multi-view green screen sequence to evaluate our reconstruction on reference views and multiple in-the-wild captures using a single camera with a resolution of $1920 \times 1080$ for every subject.
Apart from a public available sequence \textit{S4}~\cite{habermann2020deepcap}, we additionally acquired two training sequences and templates, \textit{F1} and \textit{F2}, with 18 cameras at a resolution of $1285 \times 940$, where each sequence contains around 20,000 frames. 
We will release the dataset for future research.
%
%
%
%
\vspace{-10pt}
\paragraph{Qualitative Results.}
In Fig.~\ref{fig:our_results}, we test our method on various in-the-wild environments while the subjects perform a wide range of motions. 
Our method does not only provide accurate image overlays and plausible 3D body and cloth geometries but our reconstruction also show physics-aware cloth deformations and plausible body-cloth interactions.
\textit{PADefNet} predicts different physically plausible wrinkle patterns related to the character motion as shown in Fig.~\ref{fig:dynamic_wrinkle}.
This is due to our separate modeling of body and cloth geometry and the fact that body-cloth interactions are taken into account by our simulation supervision.
We further visualize the underlying body geometry without clothing where also the occluded parts are predicted accurately.
%
%
%
%
%
\subsection{Comparisons}
We compare our approach to the state-of-the-art template-based monocular human performance capture methods~\cite{habermann2019livecap, habermann2020deepcap}.
LiveCap~\cite{habermann2019livecap} optimizes the pose via inverse kinematics to match predicted 2D and 3D joint positions and computes surface deformations via analysis by synthesis.
DeepCap~\cite{habermann2020deepcap} uses weak supervision from multi-view images during training to predict pose and embedded deformation parameters from a single segmented image.
%
%
\subsubsection{Qualitative Comparisons}
\begin{figure}[t]
    \begin{center}
    \includegraphics[width=0.11\textwidth]{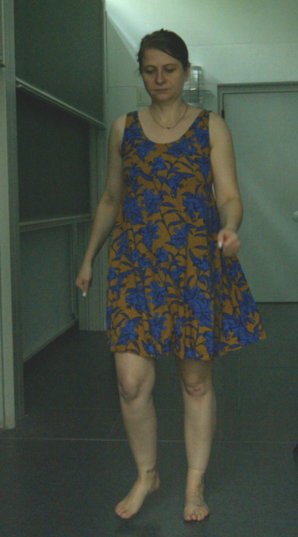}
    \includegraphics[width=0.11\textwidth]{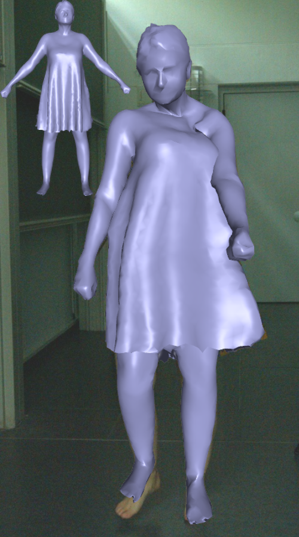}
    \includegraphics[width=0.11\textwidth]{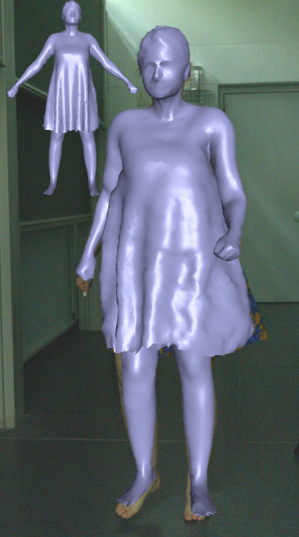}
    \includegraphics[width=0.11\textwidth]{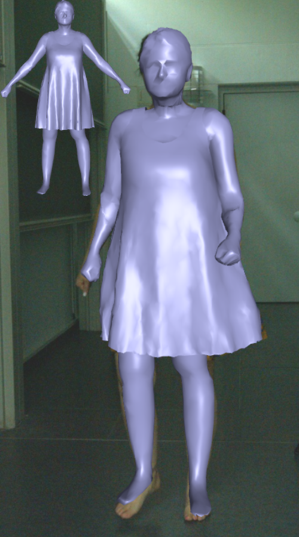}
	\bmethodlabel{0.01cm}{Input}
	\bmethodlabel{1cm}{LiveCap~\cite{habermann2019livecap}}
	\bmethodlabel{0.7cm}{DeepCap~\cite{habermann2020deepcap}}
	\bmethodlabel{0.8cm}{\textbf{Ours}}
    \end{center}
    \caption{
    Comparison with state-of-the-art template-based monocular methods~\cite{habermann2019livecap,habermann2020deepcap}. 
    Using simulation supervision during training, our method produces physically more realistic results without baked-in wrinkles from the initial template mesh (shown in the top left corners).
    }
    \label{fig:comparison}
    \vspace{-2mm}
\end{figure}

In Fig.~\ref{fig:comparison}, we compare our method with state-of-the-art template-based methods~\cite{habermann2019livecap,habermann2020deepcap}.
Unlike our approach, both LiveCap and DeepCap only use geometric priors on the deformations during optimization and training, respectively.
Consequently, the resulting cloth deformation contains static wrinkles from the initial template (see top left corners) that persist across all poses. 
By using simulation supervision and separate modeling of cloth and body geometries, the wrinkles generated by our method are less constrained by the template and, consequently, exhibit more variety and better physical plausibility.
%
%
%
%
%
\subsubsection{Quantitative Comparisons}
We evaluate our results using the green screen testing sequence of \textit{S4} for all metrics below.
Note that obtaining accurate ground truth 3D geometry from such a sparse camera system is impossible and thus we resort to image-based and physics-based metrics.
For a fair comparison, we use the same cloth-body geometry, obtained through manual clean-up of the input scans, for all approaches.
%
%
\paragraph{Out-of-balance Force Evaluation.}
In Tab.~\ref{tab:oob_force}, we list the magnitude of the out-of-balance forces, which are defined as the difference between inertial forces and the sum over internal, external, and collision forces. 
This physical measure indicates to what extent the results deviate from Newton's second law of motion and vanishes for physically correct motion.
The acceleration of the body and the garment for a given frame is determined using a centered difference approximation based on network predictions for three consecutive frames. 
To reduce the global translation error irrespective of our network predictions, we apply the ground truth global translation for all methods as described by Habermann \etal~\cite{habermann2020deepcap}.
Our method performs not only better on average compared to other approaches but also significantly reduces the peak value.
LiveCap~\cite{habermann2019livecap} performs significantly worse due to the inherent ambiguity of the single-image setting combined with the inability of geometric priors to capture physical behavior. As a result, the cloth geometry returned by their method exhibits large distortions, in particular in regions occluded from view, resulting in large internal forces.
DeepCap~\cite{habermann2020deepcap} leverages neural network models trained with multi-view supervision. 
Despite substantial improvements compared to LiveCap, our physics-aware method leads to a 50\% decrease in error.
%
%
\begin{table}[t]
\begin{center}
\begin{tabular}{ |p{2.5cm}|p{2.2cm}|p{2.2cm}|}
 \hline
 Methods     & Avg & Max\\
 \hline
 LiveCap\cite{habermann2019livecap} & 79.85 &1140\\
 \hline
 DeepCap\cite{habermann2020deepcap}   &2.119& 29.84\\
 \hline
 Ours   & \textbf{1.017} & \textbf{7.063} \\
 \hline
\end{tabular}
\end{center}
\caption{
Out-of-balance force evaluation. 
We compare our method to LiveCap~\cite{habermann2019livecap} and DeepCap~\cite{habermann2020deepcap} with respect to out-of-balance force magnitude. 
It can be seen that our physics-aware method outperforms state-of-the-art geometry-based methods.
}
\label{tab:oob_force}
\vspace{-4mm}
\end{table}
%
%
\paragraph{Garment-body Intersection Distance.}
In Tab.~\ref{tab:ixn_dis}, we further compare the garment-body intersection distance to LiveCap~\cite{habermann2019livecap} and DeepCap~\cite{habermann2020deepcap}. 
Their simplified one-piece templates sidestep this issue, however, when we evaluate with our more accurate mesh model, \eg separate geometry for clothing and body, collisions severely affect the reconstruction quality. 
We show that our network predictions significantly reduce cloth penetration.
%
%
\begin{table}[t]
\begin{center}
\begin{tabular}{ |p{4.5cm}|p{3cm}| }
 \hline
 \multicolumn{2}{|c|}{ Average Penetration Depth (cm)} \\
 \hline
 Methods     & Distance \\
 \hline
 LiveCap\cite{habermann2019livecap}  & 25.58 \\
 \hline
 DeepCap\cite{habermann2020deepcap}  & 23.83 \\
 \hline
 Ours   & \textbf{4.165} \\
 \hline
\end{tabular}
\end{center}
\caption{Penetration Depth. 
    We compute average penetration depths for cloth-body intersections across a 10,000 frame testing sequence. Not taking collisions into account, both LiveCap~\cite{habermann2019livecap} and DeepCap~\cite{habermann2020deepcap} produce severe penetrations. 
    Our method handles collisions during training, which leads to substantially reduced penetration depths. 
    }
\label{tab:ixn_dis}
\end{table}

%
%
%
%
\paragraph{IoU Percentage.}
To measure reconstruction quality from different camera views, we compare our results with previous methods using the intersection over union (IoU) metric (see Tab.~\ref{tab:iou_loss}).
The IoU metric indicates the overlapping percentage of the camera projection images of our reconstruction and the foreground segmentation of input images (ground truth).
To be consistent with DeepCap~\cite{habermann2020deepcap}, the evaluation is performed for every 100th frame of the testing sequence from \textit{S4}, and we apply the same ground truth global translation and a temporal filter.
Comparing to body-only reconstruction methods, our method achieves significantly better performance. 
It should be noted that, compared to the other approaches, the cloth geometry in our method is more constrained due to physics. For example, the strap of a dress cannot detach from the body to match the image silhouettes. 
Nonetheless, we achieve comparable IoU accuracy while maintaining better physical plausibility.
%
%
\begin{table}[t]
\begin{center}
\begin{tabular}{|c|c|c|c|}
 \hline
 Methods     & AMVIoU (\%) & RVIoU(\%) & SVIoU(\%)\\
 \hline
 HMR\cite{hmrKanazawa17} & 65.10 & 64.66 & 70.84\\
 \hline
 LiveCap\cite{habermann2019livecap} & 59.96 &59.02 &72.16\\
 \hline
 DeepCap\cite{habermann2020deepcap}   & 82.53& 82.22 & 86.66\\
 \hline
 Ours   & 80.83 & 80.53 & 84.83\\
 \hline
\end{tabular}
\end{center}
\caption{IoU percentage comparison. 
Average multi view (\textit{AMVIoU}), reference view (\textit{RVIoU}) and single view (\textit{SVIoU}) values correspond to IoU evaluation on all views, all views expect input view, and input view, respectively. 
Our reconstruction provides comparable accuracy with state-of-the-art methods while delivering more physically plausible results.}
\label{tab:iou_loss}
\end{table}
%
%
%
\subsection{Ablation Study}
\begin{figure}[ht]
    \centering
    \includegraphics[width=\linewidth]{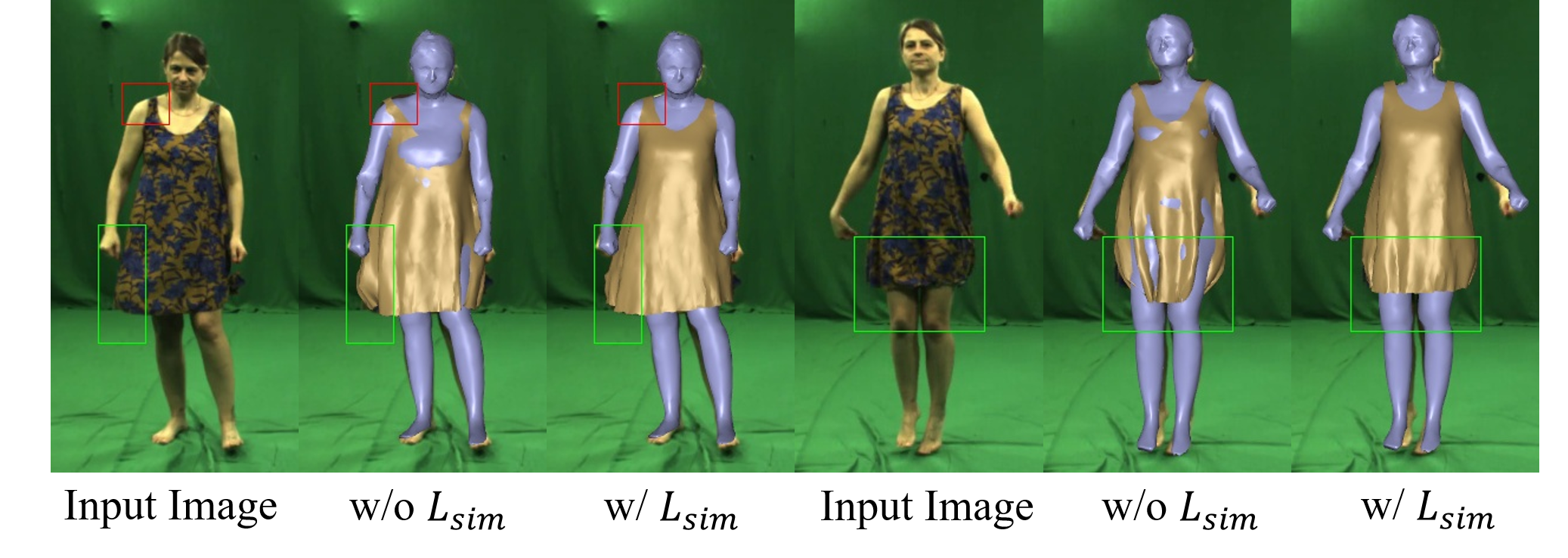}
    \caption{
    Simulation \textit{during} training. 
    We evaluate our deformation network with and without simulation loss on a testing sequence. 
    It can be seen that penetrations and deformation artifacts are largely reduced when using simulation.
    }
    \label{fig:training_with_sim}
    \vspace{-3mm}
\end{figure}

\begin{figure}[t]
    \centering
    \includegraphics[width=\linewidth]{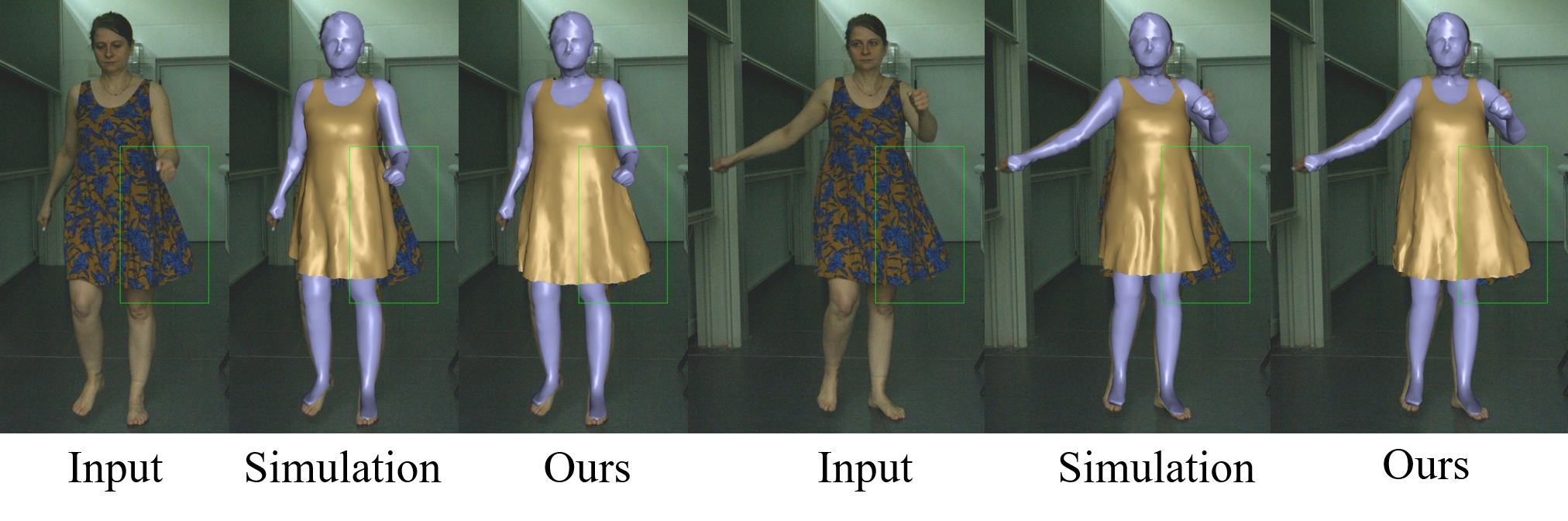}
    \caption{
    Simulation during testing. 
    Although running simulation directly on the test sequences can lead to stronger wrinkle patterns, these deformations neither match the wrinkles in the image nor the image silhouettes. 
    In contrast, our predicted cloth closely matches the silhouettes while also showing plausible deformations.
    }
    \label{fig:testing_with_sim}
\end{figure}

%
%
\paragraph{Simulation \textit{during} Training.}
Here, we verify that, with simulation supervision in the training process, physically unrealistic cloth deformations and other artifacts resulting from merely image-based supervision can be reduced.
In Fig.~\ref{fig:training_with_sim}, the strap of the dress remains on the body, and the bottom of the dress does not distort to match the silhouette. 
%
%
\vspace{-10pt}
%
\paragraph{Simulation during Testing.}
We compare our learned cloth deformation with traditional cloth simulation (TCS)~\cite{narain2012adaptive,thomaszewski2008asynchronous,Martin10EBEM} performing sequentially on a test sequence using \textit{PoseNet} to drive the body mesh.
As shown in Fig.~\ref{fig:testing_with_sim}, our results faithfully match the image evidence due to the silhouette constraint.
In contrast, TCS solely provides plausible cloth animation irrespective of the image observation.
Moreover, a single failure in the simulation process is fatal for TCS methods, since they cannot easily recover from such failure due to their sequential nature. 
As such, we refrained from reporting quantitative numbers for TCS methods as they failed already after several frames for the green screen evaluation sequences.
In contrast, the presented frame-based approach is robust to failure cases and can recover from bad frames by design. 

%
%
\section{Conclusion}
We propose a physics-aware deep learning-based method for monocular human performance capture. 
With physics-based simulation running on the fly as a network layer, we enforce physics plausibility to compensate for the shorthand of using only multi-view images. 
We show more visually pleasing results and much-improved physics metrics over state-of-the-art methods. 
%
%
\vspace{-10pt}
\paragraph{Limitations \& Future Work.}
The simulation layer is not differentiable in our current implementation. 
Nevertheless, a fully differentiable physics solver would improve data efficiency and it would open the door to automatic material parameter estimation from video input.
Our method is able to faithfully track body pose and cloth deformations for dynamic input motion, but it cannot produce dynamic effects from a single input image---an inherently ill-posed problem. 
To further improve physical fidelity and reconstruction quality, we would like to extend our method to regress dynamically consistent cloth motion by leveraging deep temporal architectures, which take short videos as input instead of single frames.

\section{Acknowledgments}
The authors would like to thank the anonymous reviewers for their valuable feedback, and Gereon Fox for the video narration.
The authors from MPII were supported by the ERC Consolidator Grant 4DRepLy (770784).

{\small
\bibliographystyle{ieee_fullname}
\bibliography{egbib}
}

\end{document}